\documentclass[11pt,a4paper,utf8,utf8x]{article}
\usepackage[hyperref]{acl2020}
\usepackage{times}
\usepackage{latexsym}

\usepackage[verbose=silent]{microtype}
\usepackage{arydshln}
\usepackage{amsmath}
\usepackage[utf8]{vietnam}
\aclfinalcopy % Uncomment this line for the final submission

\title{VLSP 2023 - LTER: A Summary of the Challenge on Legal Textual Entailment Recognition}

\author{Vu Tran$^{*1}$, Ha-Thanh Nguyen\thanks{\ \ \ \  Equal contributions}\hspace{0.1em} $^2$, Trung Vo$^3$, Son T. Luu$^3$, Hoang-Anh Dang$^3$, \\
\textbf{Ngoc-Cam Le$^4$,
    Thi-Thuy Le$^5$, Minh-Tien Nguyen$^6$, Truong-Son Nguyen$^7$, Le-Minh Nguyen$^3$}   \\
  $^1$The Institute of Statistical Mathematics, Japan \\
  $^2$National Institute of Informatics, Japan  \\
  $^3$Japan Advanced Institute of Science and Technology, Japan \\
  $^4$Vietnam Judicial Academy, Hanoi, Vietnam\\
$^5$Hanoi Law University, Hanoi, Vietnam\\
$^6$Hung Yen University of Technology and Education, Vietnam \\
  $^7$Ho Chi Minh City University of Science, Vietnam \\
}

\date{}

\begin{document}
\maketitle
\begin{abstract}
 
In this new era of rapid AI development, especially in language processing, the demand for AI in the legal domain is increasingly critical. In the context where research in other languages such as English, Japanese, and Chinese has been well-established, we introduce the first fundamental research for the Vietnamese language in the legal domain: legal textual entailment recognition through the Vietnamese Language and Speech Processing workshop. In analyzing participants' results, we discuss certain linguistic aspects critical in the legal domain that pose challenges that need to be addressed.     
\end{abstract}

\section{Introduction}

With the rapid development of AI, especially in natural language processing (NLP) tasks with the recent emergence of large language models (LLMs) and their products including, for example,  ChatGPT by OpenAI\footnote{\url{https://chat.openai.com}}, Bard by Google\footnote{\url{https://bard.google.com/}}~\cite{min2023recent,zhao2023survey}, the demand for applying AI in legal text analysis and processing is increasingly critical. Research on NLP for languages such as English, Japanese, and Chinese has been well-established. In this context, through the VLSP (Vietnamese Language and Speech Processing) workshop\footnote{\url{https://vlsp.org.vn/}}, we introduce the first fundamental research for the Vietnamese language in the legal domain\footnote{\url{https://vlsp.org.vn/vlsp2023/eval/lter}}.

The research aims to determine the legal relationship between a legal statement and a legal passage, which is fundamental to Legal AI tasks. Therefore, in this VLSP event, we will explore applications of NLP, deep learning, and generative AI for detecting the relationship between a long legal passage and a quoted statement.

Recognizing Textual Entailment (RTE) is a fundamental task in Natural Language Understanding to decide whether the meaning of a text can be inferred from the meaning of another one~\cite{dagan2022recognizing}. With that understanding, generally speaking, a legal textual entailment recognition task is to check whether a given statement is entailed by the relevant legal passage(s). The task can be described as follows:  Given a set of statements (assume S is a statement) and a set of legal passages (L$_1$, L$_2$, ..., L$_\text{N}$),  the task is required to check whether the set of legal passages entails statement~S.

With the spirit of fostering open research, participating teams may use any public resources available for the research community, for example, online law libraries like Vietnam national database of legal normative documents\footnote{\url{https://vbpl.vn}}, open-weight LLMs like Llama~2~\cite{touvron2023llama}. However, the use of closed and proprietary services (ChatGPT and its superior GPT-4, etc.) is prohibited. Results obtained through violation of this restriction are disregarded from team ranking consideration.

In this trendy research area of AI and law, other related competitions have been organized, for example, ALQAC~\cite{nguyen2023summary} focusing on information retrieval and question answering in Vietnamese law, COLIEE~\cite{goebel2023summary} focusing on information retrieval and textual entailment in Japanese and Canadian laws. Apart from them, this competition focuses on the development of fundamental research for Vietnamese legal textual entailment recognition. 

\section{Dataset}

% We prepared 76 examples for training and 139 examples for testing which cover 18 law documents. As can be seen from Table~\ref{tab:laws}, we separate the training and test data by law documents. By doing so, we can ensure the separation of training and test domains. Therefore, for systems to have high performance, the systems should be able to utilize basic language reasoning, e.g., from pretraining, and can adapt to the reasoning in the legal domain using the given training examples, but not to remember the answers from similar examples in the same domain. This will be accesses by accuracy on system predictions covering the 13 law documents not in the given the training data of labeled answers covering only 5 law documents, which can show the capacity of adaptation of the systems. With the similarity of the statistics including statement length and label balance (Table~\ref{tab:data-stats}), systems can have consistency in handling various context length and label balance by learning from the training data.  

We prepared 76 examples for training and 139 examples for testing which cover 18 law documents as can be seen from Table~\ref{tab:laws}, the majority of the law documents in the test dataset are not shared or available in the training dataset. By doing so, we can have some separation between training and test domains. Therefore, for systems to have high performance, the systems should be able to utilize basic language reasoning, e.g., from pretraining, and can adapt to the reasoning in the legal domain using the given training examples, but not only to remember the answers from similar examples in the same domain. This will be assessed by accuracy on system predictions covering the 13 law documents not in the given training data of labeled answers covering only 5 law documents, which can show the capacity of adaptation of the systems. With the similarity of the statistics, including statement length and label balance (Table~\ref{tab:data-stats}), systems can have consistency in handling various context length and label balance by learning from the training data.

\begin{table*}[t]
    \centering
    
    \begin{tabular}{|p{5cm}|p{7cm}|r|r|}
    \hline
        \textbf{Law Document} & \textbf{English Translation} & \textbf{Train} & \textbf{Test} \\ \hline
        Hiến pháp 2013 & Constitution 2013 & 0 & 9 \\
Bộ Luật Dân sự 2015 & Civil Code Law 2015 & 0 & 18 \\
Luật Phòng, chống bạo lực gia đình 2022 & Law on Domestic Violence Prevention And Control 2022 & 0 & 11 \\
Luật Thanh tra 2022 & Law on Inspection 2022 & 0 & 20 \\
Luật Tố tụng hành chính 2015 & Administrative Procedure Law 2015 & 0 & 3 \\
Luật Hôn nhân và gia đình 2014 & Law on Marriage and Family 2014 & 0 & 11 \\
Luật Dầu khí 2022 & Petroleum Law 2022 & 0 & 9 \\
Luật Trọng tài thương mại 2010 & Commercial Arbitration Law 2010 & 0 & 8 \\
Luật Cư trú 2020 & Law on Residence 2020 & 0 & 8 \\
Luật Tiếp cận thông tin 2016 & Law on Access to Information 2016 & 0 & 5 \\
Luật An ninh mạng 2018 & Law on Cybersecurity 2018 & 0 & 2 \\
Luật Du lịch 2017 & Law on Tourism 2017 & 0 & 1 \\ 
Luật Tổ chức viện kiểm sát nhân dân 2014 & Law on the Organization of the People's Procuracy 2014 & 0 & 9 \\ \cdashline{1-4}
Luật Giáo dục 2019 & Education Law 2019 & 21 & 4 \\
Luật Điện ảnh 2022 & Film Law 2022 & 7 & 8 \\
Luật Viên chức 2010 & Law on Public Employees 2010 & 21 & 4 \\
Luật Thanh niên 2020 & Law on Youth 2020 & 8 & 3 \\
Luật Phòng, chống ma túy 2021 & Law on Prevention and Control of Narcotic Substances 2021 & 19 & 6 \\

% Luật Điện ảnh 2022 & Film Law 2022 & 7 & 0 \\
% Luật Phòng, chống ma túy 2021 & Law on Prevention and Control of Narcotic Substances 2021 & 19 & 0 \\
% Luật Viên chức 2010 & Law on Public Employees 2010 & 21 & 0 \\
% Luật Giáo dục 2019 & Education Law 2019 & 21 & 0 \\
% Luật Thanh niên 2020 & Law on Youth 2020 & 8 & 0 \\
\hline
    \end{tabular}
    \caption{Number of questions for each law document.}
    \label{tab:laws}
\end{table*}

\begin{table}[ht]
    \centering
    \begin{tabular}{l|r|r}
       \textbf{Stats}  & \textbf{Train} & \textbf{Test} \\ \hline
        \#Examples & 76 & 139 \\
        Average length (\#syllables) & 30 & 27 \\
        Min length & 8 & 7 \\
        Max length & 76 & 72 \\
        \%Yes labels & 54\% & 47\% \\
    \end{tabular}
    \caption{Data statistics: number of examples, statement's syllable counts, and label balance.}
    \label{tab:data-stats}
\end{table}

\begin{figure}[ht!]
    \centering
\fbox{\parbox{0.475\textwidth}{
[

    \hspace{0.5cm}\{
    
        \hspace{0.5cm}\hspace{0.5cm}"example\_id": "DS-101",
        
        \hspace{0.5cm}\hspace{0.5cm}"label": "Yes/No",
        
        \hspace{0.5cm}\hspace{0.5cm}"statement": "Cơ sở điện ảnh phát hành phim phải chịu trách nhiệm trước pháp luật về nội dung phim phát hành là sai.",
        
        \hspace{0.5cm}\hspace{0.5cm}"legal\_passages": [
            
            \hspace{0.5cm}\hspace{0.5cm}\hspace{0.5cm}\{
            
                \hspace{0.5cm}\hspace{0.5cm}\hspace{0.5cm}\hspace{0.5cm}"type": "law",
                
                \hspace{0.5cm}\hspace{0.5cm}\hspace{0.5cm}\hspace{0.5cm}"law\_id": "05/2022/QH15",
                
                \hspace{0.5cm}\hspace{0.5cm}\hspace{0.5cm}\hspace{0.5cm}"article\_id": "15"
                
            \hspace{0.5cm}\hspace{0.5cm}\hspace{0.5cm}\}
            
        \hspace{0.5cm}\hspace{0.5cm}]
        
    \hspace{0.5cm}\}
    
]
}}

(Annotated data)

\vspace{0.5cm}

\fbox{
\parbox{0.475\textwidth}{

[

\hspace{0.5cm}\{

\hspace{0.5cm}\hspace{0.5cm}"id": "05/2022/QH15",

\hspace{0.5cm}\hspace{0.5cm}"articles": [

\hspace{0.5cm}\hspace{0.5cm}\hspace{0.5cm}\{

\hspace{0.5cm}\hspace{0.5cm}\hspace{0.5cm}\hspace{0.5cm}"id": "15",

\hspace{0.5cm}\hspace{0.5cm}\hspace{0.5cm}\hspace{0.5cm}"text": " ... "

\hspace{0.5cm}\hspace{0.5cm}\hspace{0.5cm}\}

\hspace{0.5cm}\hspace{0.5cm}]

\hspace{0.5cm}\}

]
}}

(Provided legal passages)

    \caption{Data format}
    \label{fig:data-format}
\end{figure}

As a test for the consistency of logical reasoning, we design examples that are related or similar to each other. For instance, a) "Nguyễn Văn A sinh ngày 03/10/2003 thì ngày 02/10/2023 A đủ tuổi kết hôn" and b) "Nguyễn Văn A sinh ngày 03/10/2003 thì ngày 03/10/2023 A đủ tuổi kết hôn". A system with the capability of consistent logical reasoning should not answer incorrectly any examples in a related group. We will see to what extent the participants achieve this capability in the Evaluation section.

We provided participants training/test data and required legal passages in JSON format as shown in Figure~\ref{fig:data-format}.

\section{Participants' Approaches}
Out of the 22 registered teams, the five final teams that stand out include CAN NOT STOP, NOWJ, A3N1, Angels, and HNO3. Interestingly, all five of these teams employ large language models (LLMs) in their approaches, signifying the growing importance and prevalence of LLMs in tasks related to semantic analysis in Vietnamese legal text.

\paragraph{CAN NOT STOP}\cite{vlsplter2023cannotstop} Their framework utilizes label models to ensemble predictions from two available large language models (LLMs). The first LLM is mT0, a multilingual LLM that is instruction-finetuned on a mixture of tasks. The second LLM is a Vietnamese Llama-2 model that is finetuned specifically for the task of legal text entailment. They prompt the two LLMs with a temperature variable greater than zero to obtain many provisional (often noisy) predictions. Afterward, they leverage label models, which help adjudicate predictions based on agreements and disagreements among them, to ensemble the noisy predictions.

\paragraph{NOWJ}\cite{vlsplter2023nowj} They proposed methods for this task involving three
phases: translation, clause matching, and utilization of large language models. The translation process converts Vietnamese inputs into English inputs to utilize large language models excelling at processing English. Then, the clause-matching process starts with implementing regex techniques to extract clauses and their corresponding titles from the given articles. Then, BM25 Okapi is applied to calculate the relevant score of each clause which is considered relevant if its score is greater than a chosen threshold by parameter search. Finally, each statement and its relevant clauses are put into a designated template fed to an LLM, which emits a response indicating Yes or No. 

\paragraph{A3N1}\cite{vlsplter2023a3n1}  They proposed on top of a pre-trained language model a layer for fine-tuning based on Support Vector Machines~\cite{cortes1995support} kernel
formulations. The proposed method leverages an intelligent
combination of two popular kernel functions, polynomial kernel and Gaussian kernel, tightly integrated with the last layer of the pre-trained language model. They call this layer the ``SVM layer'' because
it provides an innovative computational formula
that efficiently processes data and enhances the representational capabilities of the model. 

\paragraph{Angels}\cite{vlsplter2023angels} They adopt a transfer learning approach, focusing on transformer-based language models~\cite{vaswani2017attention} for semantic textual similarity tasks in Vietnamese legal text. They explore a range of pre-trained models applicable to Vietnamese. To enhance the models’ understanding of nuanced semantic relationships, they employ a contrastive learning approach. Specifically, they utilize the CrossEncoder architecture. The CrossEncoder processes pairs of texts simultaneously in a single forward pass, allowing it to capture intricate interactions between the claim and sentence comprehensively.

\paragraph{HNO3}\cite{vlsplter2023hno3} They utilized a pre-trained language model  BERT (Bidirectional Encoder Representations from Transformers) specifically for Vietnam Legal Law (VNBert Law). Due to the limited training data, they added additional data from ALQAC 2021 and broke down the legal articles into smaller samples based on their individual items. To evaluate and label the data, they split the "legal\_passage" into sentences that had
similar similarity ratings as determined by the BM25\_Score with the corresponding
statements. The pairs with the highest BM25\_Score were considered as representing the
best match between the "legal\_passage" and the "statement".
The input format for the BERT-based model consisted of the following structure: [CLS] Statement [SEP] BM25\_Score [SEP] Legal\_passage [SEP].
\section{Evaluation}

The evaluation measure is accuracy with respect to whether the Yes/No label was correctly confirmed. 

\begin{table}[ht]
    \centering
    \begin{tabular}{l|c}
        \textbf{Team Name} &  \textbf{Accuracy} \\ \hline
        CAN NOT STOP  & \textbf{0.7698} \\
        NOWJ  & 0.7626 \\
        A3N1 & 0.7194 \\
        Angels & 0.5468 \\
        HNO3  & 0.5324
    \end{tabular}
    \caption{Official results of the participants.}
    \label{tab:results}
\end{table}

\paragraph{Logical consistency} It is to check if the system can have consistent predictions for related examples. Related examples are grouped by their (very high) similarity, for instance, a) "Nguyễn Văn A sinh ngày 03/10/2003 thì ngày 02/10/2023 A đủ tuổi kết hôn" and b) "Nguyễn Văn A sinh ngày 03/10/2003 thì ngày 03/10/2023 A đủ tuổi kết hôn". As shown in Tables~\ref{tab:results-consistency-reduction}, when considering this, the accuracy drops substantially, which indicates that participants' systems have difficulties in logically reasoning of the details. Among those participants, team NOWJ has a system with the best logical consistency (Tables~\ref{tab:results-consistency-reduction}~and~\ref{tab:results-consistency}). 

\begin{table}[ht]
    \centering
    \begin{tabular}{l|c}
        \textbf{Team Name} &  \textbf{Accuracy} (reduction) \\ \hline
        CAN NOT STOP  & 0.5899 (-0.1799) \\
        NOWJ  & \textbf{0.6978} (\textbf{-0.0647})\\
        A3N1 & 0.5755 (-0.1439)\\
        Angels & 0.3237 (-0.2230)\\
        HNO3  & 0.3381 (-0.1942)
    \end{tabular}
    \caption{Performance of the participants when considering that all related examples must be correctly answered. These results can be compared with ones in Table~\ref{tab:results} for assessing performance degradation.}
    \label{tab:results-consistency-reduction}
\end{table}

\begin{table}[ht]
    \centering
    \begin{tabular}{l|c}
        \textbf{Team Name} &  \textbf{Accuracy} \\ \hline
        CAN NOT STOP  & 0.5313  \\
        NOWJ  & \textbf{0.7188}\\
        A3N1 & 0.5625 \\
        Angels & 0.2188 \\
        HNO3  & 0.1563
    \end{tabular}
    \caption{Performance of the participants on the subset of all related examples grouped by relevance where a group prediction is considered accurate if and only if all of its examples must be correctly answered. Accuracy is, thus, measured based on the number of groups accurately predicted. No. groups:~32.}
    \label{tab:results-consistency}
\end{table}

\paragraph{Negation} It is one of the fundamental problems in natural language processing. If a system is built with content-similarity in mind, e.g., word co-occurrence frequency, a negated sentence is highly similar to its origin, thus indistinguishable by such a system. To deal with the negation problem, a system needs to have high logical consistency. To see this, we measure the accuracy of the systems on the subset of the test examples where we have pairs of a statement and its paired negated statement, where a system must provide correct answers for both statements of a pair to score. The results are shown in Table~\ref{tab:results-negation}. The top-3 teams of overall accuracy also have the best performance on negation-pairs. Though, the accuracy in this assessment is only 0.6154, which means there is much more room for improvement. Even systems utilizing LLMs may still suffer from their limitations in dealing with negation~\cite{nguyen2023negation}.

\begin{table}[ht]
    \centering
    \begin{tabular}{l|c}
        \textbf{Team Name} &  \textbf{Accuracy} \\ \hline
        CAN NOT STOP  & \textbf{0.6154} \\
        NOWJ  & \textbf{0.6154} \\
        A3N1 & \textbf{0.6154} \\
        Angels & 0.0000 \\
        HNO3  & 0.0769 
    \end{tabular}
    \caption{Performance of the participants when measured on the subset of the similar examples that have a negated counterpart. No. negation-pairs:~13. }
    \label{tab:results-negation}
\end{table}

\paragraph{Difficulty of the task} Over 139 test examples, 29/139 examples were predicted correctly by all 5 teams, 1/139 example was predicted incorrectly by all 5 teams, and 107 examples were correctly predicted by at least 3/5 teams. When considering the logical consistency problem, this legal textual entailment recognition task is considerably challenging, where 21/139 examples were predicted correctly by all 5 teams, 12/139 example was predicted incorrectly by all 5 teams, and 71 examples were correctly predicted by at least 3/5 teams. Even though our dataset is created essentially based on the main text of the law documents, in concrete legal events, things, and their relationships are much more complicated, so there is still a lot of work to be done on the way, especially to deal with inherent challenges in utilizing LLMs for legal text processing~\cite{anh2023impact} including, for instance, interpretability, explainability, and bias mitigation.

\section{Conclusion}

We have presented a summary of the first held challenge for legal textual entailment recognition in the Vietnamese language through VLSP. Through the analysis of participants' results, we first observed that 1)the emergence of LLMs which can be already applicable in the legal domain in Vietnamese with potential performance, 2) despite this, the ability of the systems to pay attention to details and especially negation is limited, thus, needs to be improved much more. In the future direction of this challenge, we expect to tackle more fine-grained linguistic aspects critical to legal textual entailment recognition.  

\section*{Acknowledgments}
We would like to express our deep gratitude to VLSP's general organizers for organizing the workshop where we can together tackle this LTER challenge, sponsors and supporters for their continuous support much needed for holding up the challenge, and participants for their contributions of research ideas and implementations which are much important for pushing the research forwards.    
% \cite{vlsplter2023summary}

\bibliography{acl2020}
% \bibliography{anthology,acl2020}
\bibliographystyle{acl_natbib}

\end{document}